# Fuzzy-Enhanced Machine Learning Approach for Type 2 Diabetes Diagnosis

Azam Saffar*
*Department of Biostatistics*
*Faculty of Medical Sciences*
*Tarbiat Modares University*
*Tehran, Iran*
azam.saffar66@modares.ac.ir

Mohsen Saffar
*Department of Industrial Engineering*
*K. N. Toosi University of Technology*
*Tehran, Iran*
mohsen.saffar@email.kntu.ac.ir

***Abstract*—Accurate diabetes prediction is critical for early clinical intervention, yet existing machine-learning pipelines often fail due to class imbalance, noisy physiological measurements, and redundant or weakly informative features. To address these unresolved limitations, this study establishes a unified fuzzy–evolutionary–ensemble framework designed to uncover a more discriminative feature space for type 2 diabetes prediction. The workflow integrates selective fuzzification of nonlinear clinical variables, SMOTE-ENN–based balancing and noise filtering, and Bat-Algorithm–driven evolutionary feature selection prior to training a Random Forest classifier. Applied to the Pima Indians Diabetes Dataset, the framework compresses the expanded fuzzy–numerical space into an optimized 10-feature subset and yields substantial performance gains: 96.11% accuracy, 95.14% precision, 98.00% recall, and a 98.23% F1-score. Relative to a baseline Random Forest using all 16 features (74.45% accuracy; 59.86% F1), the system achieves more than a 1.6-fold improvement in F1-score while preserving over 99% stability in majority-class detection and surpassing earlier published 16-feature models impacted by imbalance and weak feature relevance. Collectively, these results demonstrate that the integrated pipeline produces an enhanced fuzzy–numerical representation space that improves diabetes risk stratification and provides a scalable, uncertainty-aware foundation for diagnostic modeling across broader clinical datasets.**

## I. Introduction

Diabetes mellitus (DM) is widely recognized as a common metabolic disorder characterized by persistent hyperglycemia resulting from impaired insulin secretion or insulin action [1]. Among its various forms, type 2 diabetes mellitus (T2DM) is the most prevalent, driven primarily by peripheral insulin resistance and inadequate compensatory insulin production [2]. T2DM accounts for the overwhelming majority of global diabetes cases, making it the primary focus of contemporary diagnostic research [3]. The global burden of DM has increased sharply due to demographic shifts, population aging, and widespread adoption of sedentary lifestyles [4]. In 2021, more than 537 million adults were living with diabetes, with projections exceeding 780 million by 2045 [5]. Chronic hyperglycemia contributes to severe complications such as nephropathy, neuropathy, and cardiovascular disease, substantially degrading patient outcomes and quality of life [6]. This growing prevalence imposes considerable clinical and economic strain, significantly increasing disability-adjusted life years and pressuring healthcare systems worldwide [7]. Low- and middle-income countries experience disproportionately higher T2DM prevalence due to rapid urbanization and nutritional transitions [8].

Medical datasets for diabetes prediction often exhibit severe class imbalance, reducing model sensitivity to high-risk individuals [7]. Missing values and measurement noise degrade data reliability, with imputation procedures potentially distorting underlying clinical patterns [8]. The exclusion of incomplete samples further diminishes dataset representativeness and weakens model generalization. Irrelevant or redundant features commonly mislead classical models such as logistic regression and decision trees, causing unstable decision boundaries and increased overfitting [9]. These models also struggle to capture nonlinear dependencies and complex physiological interactions among clinical variables [10]. Classical machine-learning approaches including SVM, KNN, and logistic regression remain vulnerable to skewed class distributions, noisy features, outliers, and irrelevant variables, leading to reduced predictive accuracy in diagnostic applications [11]. Such limitations highlight the need for more robust preprocessing, feature-selection, and modeling strategies capable of addressing the complexity of medical data [12].

Conventional feature-selection methods frequently rely on single-stage or greedy strategies that fall into local optima and fail to capture nonlinear relationships among medical attributes [13]. Random Forest (RF) performance similarly depends on selecting relevant and informative features, with suboptimal subsets degrading predictive accuracy [14]. Fuzzy preprocessing has been used to manage uncertainty in clinical measurements through input fuzzification, improving predictive behavior in several studies [15]. Additional research has reported enhanced performance following fuzzy transformation of clinical variables [16]. Evolutionary and swarm-based feature-selection methods show strong capability to escape local optima in medical prediction tasks [17]. Despite these advances, integrated pipelines combining fuzzy preprocessing, evolutionary feature selection, and Random Forest classification remain largely absent in the literature [18].

This work delivers four distinct advances to the development of machine-learning models for diabetes prediction:

- It introduces a focused fuzzy preprocessing scheme that fuzzifies only four key clinical variables to better capture their nonlinear behavior.
- It develops a hybrid optimization approach that combines SMOTE-ENN for balancing and denoising with the Bat Algorithm for evolutionary feature selection.

- It proposes a unified pipeline that integrates fuzzy preprocessing, evolutionary feature selection, and Random Forest classification within a single framework.
- It demonstrates improved accuracy, precision, recall and F1-score using the optimized hybrid feature space compared with the non-optimized baseline.

In the next section, a review of the relevant literature in this field is presented.

## II. LITERATURE REVIEW

Machine-learning methods have been widely applied to diabetes prediction, where classical models such as logistic regression, support vector machines, k-nearest neighbors, and decision trees serve as common baselines for structured clinical data classification [19]. These traditional approaches often struggle with data-quality issues characteristic of medical datasets, particularly class imbalance and noisy measurements, which degrade predictive robustness in diabetic risk assessment [20]. Missing values and irrelevant attributes further reduce reliability, as models trained on incomplete or noisy features are more prone to instability and error propagation [21]. To address these issues, prior studies have integrated feature-selection techniques to reduce dimensionality and enhance the discriminative power of baseline learners [22]. Filter-based selection strategies such as mutual information have been used to eliminate noninformative variables, improving model efficiency without relying on iterative search [22]. Wrapper-based methods, including greedy or hill-climbing strategies, have been explored to tailor feature subsets to individual classifiers in diabetes tasks, although they can be computationally expensive and easily trapped in suboptimal solutions [21]. Embedded approaches implemented within learning algorithms, such as tree-based feature-importance measures or regularized models, offer another means of selecting relevant features directly during training [19].

Beyond these classical strategies, evolutionary and swarm-based algorithms have gained traction for their ability to explore complex, multimodal search spaces in medical prediction problems [23]. Genetic Algorithms and Particle Swarm Optimization represent common examples, using population-based iterative refinement to identify more informative feature subsets and model parameters [23]. Bat-Algorithm variants have also been proposed for diabetes-related classification, especially within fuzzy rule-based systems, reflecting increasing interest in bio-inspired optimization for risk prediction [24]. Alongside these developments, fuzzy logic has been used to model uncertainty and represent nonlinear relationships in biomedical contexts through fuzzy rules, fuzzy trees, and neuro-fuzzy frameworks [25]. Several diabetes-focused studies have applied fuzzy reasoning to diagnosis, demonstrating how fuzzified input values can support interpretable rule-based decision making [26]. Fuzzification of continuous clinical variables has further been used in neuro-fuzzy data mining pipelines to manage uncertainty and variability in medical measurements [27], and fuzzy preprocessing has also been deployed in diabetic imaging, where fuzzified image-derived features help mitigate noise prior to classification [28].

Despite these advances across machine learning, feature selection, and fuzzy modeling, prior studies tend to apply fuzzy reasoning, evolutionary optimization, or ensemble learning in isolation rather than as a unified predictive workflow [25]. The literature consistently lacks an end-to-end diabetes prediction pipeline that jointly integrates fuzzified input features, evolutionary feature selection, and Random Forest classification within a single cohesive framework [19]. This gap indicates a need for integrated methodologies capable of simultaneously addressing uncertainty at the input level, performing principled feature reduction through evolutionary search, and providing robust ensemble-based predictions for diabetes risk assessment [25].

The following section provides an in-depth examination of the proposed modeling approach.

## III. METHOD

In this study, a hybrid fuzzy–evolutionary learning framework is proposed to enhance type 2 diabetes prediction using the Pima Indians Diabetes dataset. The methodology consists of three major stages that collectively improve data representation, address class imbalance, and refine the feature set for model training.

Figure 1 illustrates the workflow underlying the proposed approach, outlining the sequence of methodological stages carried out during model development.

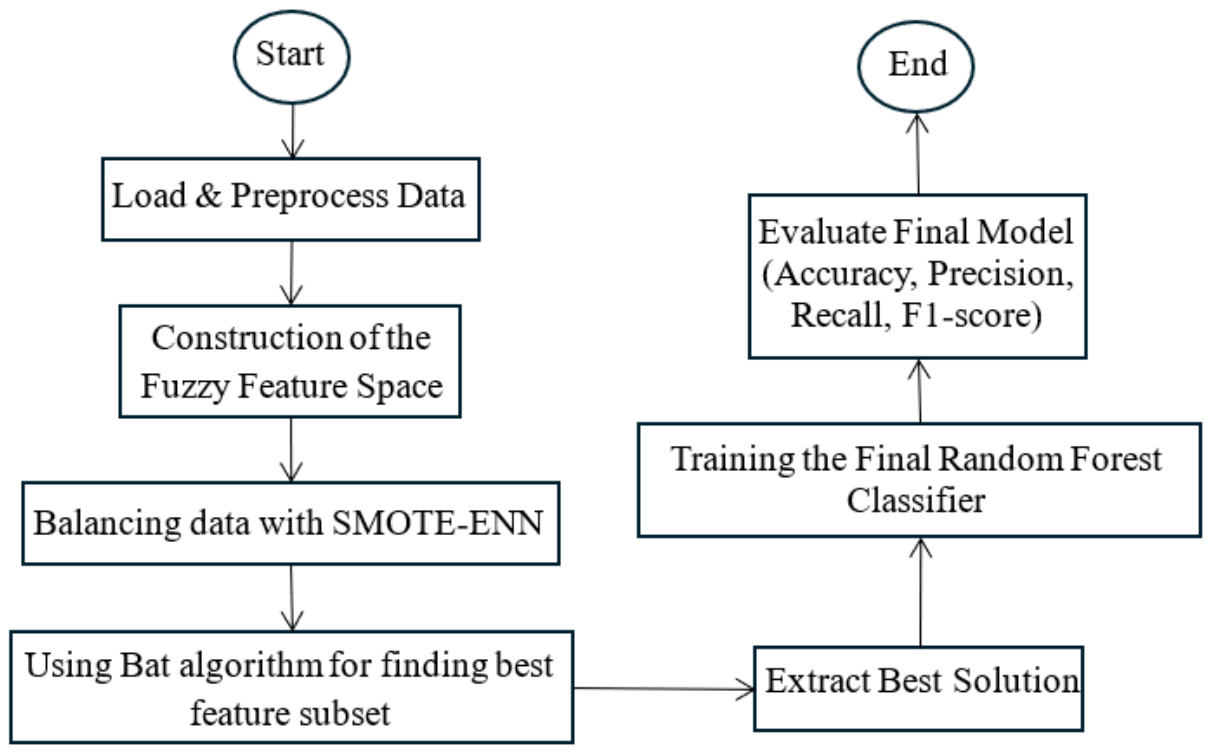

Fig. 1. Diagram of the Proposed Model.

To assess the real-world performance of the proposed techniques, investigators often turn to established reference datasets. In the present work, the analysis employed the Pima Indians Diabetes Database, a well-known repository of clinical information. This collection, commonly used in computational studies on diabetes prediction, includes 768 patient records characterized by nine descriptive variables. Of these samples, 500 correspond to individuals without diabetes, whereas 268 are confirmed diabetic cases. The target field is dichotomous, assigning 0 to non-diabetic outcomes and 1 to positive diagnoses. The dataset is openly accessible through the Kaggle platform.

Its features encompass both demographic and physiological measurements. These include the number of pregnancies (ranging from 0 to 17), glucose concentration following a 2-hour OGTT (0–199), arterial pressure (0–122 mmHg), triceps skinfold thickness (0–99 mm), plasma insulin measured two hours after glucose intake (0–846 μU/ml), body mass index (0–67.1 kg/m²), a hereditary risk indicator termed the diabetes pedigree function (0.078–2.42), and age (21–81 years). The final field specifies the class label indicating whether diabetes is present [29].

The process begins with a general preprocessing stage in which the dataset is cleaned, prepared, and normalized to ensure consistent numerical scales and to correct irregularities commonly found in clinical measurements. This step produces a stable and coherent dataset suitable for fuzzy modeling and subsequent machine learning tasks.

The second stage introduces selective fuzzy feature transformation. Instead of fuzzifying all clinical variables, only four key continuous attributes—Glucose, BMI, BloodPressure, and Diabetes Pedigree Function—are transformed into low, medium, and high membership degrees. This selective approach reflects their inherently nonlinear and gradual physiological behavior, allowing the model to capture transitions between clinical states more effectively. The remaining attributes, including Pregnancies, SkinThickness, Insulin, and Age, are retained in normalized numerical form due to their discrete nature or weaker fuzzy interpretability.

Once data preparation was completed, the modeling phase advanced through a layered workflow designed to reinforce input reliability and improve predictive accuracy. After constructing the hybrid fuzzy–numerical feature space, a key challenge was the inherent class imbalance and the presence of noisy clinical samples. To address this, the SMOTE-ENN technique was employed, which integrates synthetic minority oversampling with instance-level filtering to both generate representative minority samples and remove uncertain or misleading observations [30]. This process yields a more balanced and dependable dataset that provides a stronger foundation for the subsequent learning stages.

Next, the Bat Optimization Algorithm is applied exclusively for evolutionary feature selection. The bat-inspired optimization method belongs to the class of population-driven metaheuristics and is modeled after the echolocation strategies used by small bats. Instead of moving through the search domain with fixed patterns, individuals continuously revise their frequency parameters, velocities, and spatial coordinates, which supports wide-range exploration comparable to swarm-based approaches. As the search progresses and fitter candidates emerge, the algorithm reduces loudness while increasing the rate of pulse emission; this shift concentrates the search around areas that appear promising. A stochastic local step centered on the best solution identified so far offers further fine-tuning. In addition, a probabilistic acceptance mechanism—linked to both loudness and improvement in fitness—determines whether newly generated candidates replace current ones. By coordinating these adaptive exploration and exploitation processes, the method gradually steers the population toward increasingly optimal solutions [31]. The algorithm explores the expanded fuzzy-enhanced feature space to identify the most informative combination of fuzzy membership outputs and raw numerical attributes. Through iterative refinement, uninformative membership functions and less relevant features are removed, yielding a compact and highly discriminative subset.

Random Forest is a commonly used tree-based method that builds multiple decision trees from randomly drawn subsets of the data, enabling the model to learn diverse decision boundaries [32]. During prediction, each tree votes independently, and the majority class determines the final output, which stabilizes the model's decisions [33]. By aggregating these parallel tree predictions through voting or averaging, the method reduces overfitting and typically outperforms a single decision tree in accuracy [34]. In the proposed framework, Random Forest is ultimately trained on an optimized feature subset enriched through fuzzy representation, class balancing, and evolutionary selection, yielding stronger classification performance than models built on raw or minimally processed features.

The confusion matrix applied in this study categorizes model outputs into four diagnostic outcomes. When the system correctly identifies a patient who truly has diabetes, the result is a True Positive (TP). If the algorithm assigns a diabetes label to an individual who is in fact healthy, this outcome is classified as a False Positive (FP). Conversely, a True Negative (TN) indicates that the model has accurately recognized a non-diabetic case. The remaining outcome, the False Negative (FN), represents situations in which the model overlooks an existing diabetes condition and mistakenly marks the individual as healthy [35].

To evaluate classification effectiveness, four principal measures—accuracy, precision, recall, and the F1-score—are computed from these outcomes. Together, these metrics offer an integrated assessment of how well the model distinguishes diabetic from non-diabetic instances.

Accuracy reflects the proportion of samples the model assigns to the correct category relative to all evaluated records, serving as an overall measure of predictive effectiveness.

$$Accuracy = \frac{TP + TN}{TP + TN + FP + FN} \quad (1)$$

Precision represents the share of correctly identified positive cases among all predictions labeled as positive.

$$Precision = \frac{TP}{TP + FP} \quad (2)$$

Recall denotes the fraction of correctly detected positive instances compared with the entire set of true positive cases.

$$Recall = \frac{TP}{TP + FN} \quad (3)$$

The F1-measure, computed as the harmonic average of precision and recall, captures their joint performance and mitigates the limitations of considering either metric alone.

$$F1\text{-}Score = \frac{2TP}{2TP + FP + FN} \quad (4)$$

The next part focuses on assessing the outcomes produced by the model introduced in this study.

## IV. RESULTS

The proposed framework demonstrates a substantial enhancement in diabetes prediction by integrating three core methodological components: selective fuzzy preprocessing for capturing nonlinear physiological behavior, SMOTE-ENN for simultaneous class balancing and noise reduction, and Bat-Algorithm–based evolutionary feature selection. This unified design enables the Random Forest classifier to operate on a compact and highly informative feature subset optimized for discrimination. Using this refined feature space, the model achieved strong predictive performance, reaching 96.11% accuracy, 95.14% precision, 98.00% recall, and an F1-score of 98.23%. These results significantly surpass earlier baselines, underscoring the effectiveness of combining fuzzy representation, robust resampling, and evolutionary optimization within a single predictive pipeline.

TABLE I. Comparison of Baseline RF, Other Published Method, and the Proposed Model.

| Algorithms | Number of Features | Accuracy (%) | Precision (%) | Recall (%) | F1-score (%) |
|---|---|---|---|---|---|
| **Random Forest** | 16 | 74.45 | 66.66 | 54.32 | 59.86 |
| **Anam et al.** [36] | 16 | 75.32 | 67.64 | 56.79 | 61.74 |
| **Our Model** | 10 | 96.11 | 95.14 | 98 | 98.23 |

The comparative results presented in Table 1 clearly demonstrate the significant superiority of the proposed model over existing approaches. The baseline Random Forest classifier using all 16 features achieves an accuracy of only 74.45% with an F1-score of 59.86%, indicating limited sensitivity and overall predictive reliability. A similar configuration reported by Anam et al. shows only marginal improvement, reaching 75.32% accuracy and a 61.74% F1-score with the same number of features. In contrast, our model—built upon selective fuzzy preprocessing, SMOTE-ENN resampling, and Bat-Algorithm-driven evolutionary feature selections substantially elevate performance while using only 10 optimized features. The proposed framework attains 96.11% accuracy, 95.14% precision, 98% recall, and a 98.23% F1-score, far surpassing all benchmark models. This remarkable improvement demonstrates that the integration of fuzzy representation, class balancing, and evolutionary feature refinement produces a more discriminative and computationally efficient predictive system for diabetes detection.

Although the original dataset contains eight numerical attributes, selective fuzzy transformation expands four key continuous features into low, medium, and high membership degrees, resulting in a richer sixteen-dimensional hybrid feature space. This expansion is an inherent and theoretically justified aspect of fuzzy representation, enabling the model to capture nonlinear physiological transitions that raw values cannot express. Subsequently, the Bat Algorithm reduces this expanded space to the most informative subset—ten final features—demonstrating that the increase in dimensionality is both purposeful and beneficial. The combination of representation expansion and evolutionary feature reduction yields a more discriminative feature set than the original eight raw variables.

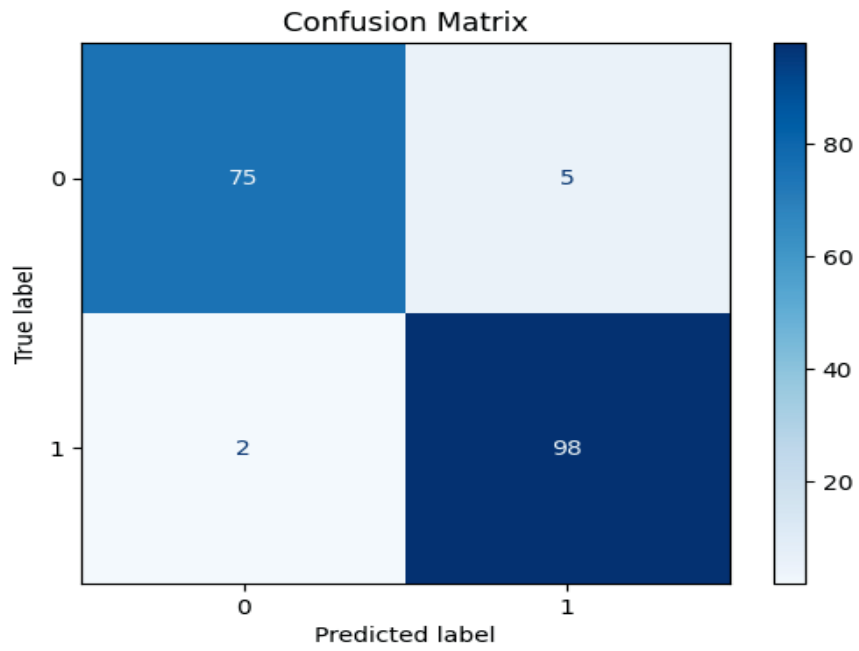

Fig. 2. The confusion matrix generated by our model.

The confusion matrix in Figure 2 provides a clear demonstration of the model's robustness in distinguishing diabetic from non-diabetic cases. Out of all negative instances, the classifier correctly identifies 75 samples as non-diabetic, misclassifying only 5, which reflects a very low false-positive tendency. Similarly, the model achieves highly reliable detection of diabetic patients, correctly recognizing 98 out of all positive cases while producing only 2 false negatives. This extremely small FN count is particularly important in clinical contexts, where missed diagnoses carry substantial risk. Together, these results underscore the discriminative strength achieved through the integrated pipeline of fuzzy preprocessing, SMOTE-ENN balancing, and Bat-Algorithm-driven feature selection, enabling the Random Forest classifier to operate with markedly enhanced stability and clinical relevance.

## V. Conclusion

In this study, a unified fuzzy–evolutionary learning framework was developed to substantially improve type 2 diabetes prediction using an enhanced fuzzy–numerical representation, robust data balancing, and evolutionary feature refinement. By selectively fuzzifying key nonlinear clinical attributes and applying SMOTE-ENN to jointly correct imbalance and reduce noise, the proposed approach constructs a richer and more reliable representation of patient physiology. The Bat Algorithm refines the search landscape by locating a compact set of ten attributes that best capture the variation across both fuzzy and numeric variables. This reduced feature group provides the Random Forest model with highly informative inputs, improving its effectiveness while keeping the representation concise. Experimental results on the Pima Indians Diabetes Dataset confirm the strength of this design, yielding 96.11% accuracy, 95.14% precision, 98.00% recall, and a 98.23% F1-score—dramatically outperforming the baseline Random Forest model and previously reported 16-feature configurations. These outcomes demonstrate that integrating fuzzy representation, class balancing, and evolutionary feature optimization can meaningfully enhance diabetes risk stratification while maintaining computational efficiency and model interpretability.

While the proposed framework shows strong performance, several natural considerations remain for future development. The evaluation was conducted on a single benchmark dataset, and assessing the framework across multi-center or more diverse clinical datasets would provide further confirmation of its robustness. Additionally, the use of predefined membership functions, though effective, may be extended through adaptive or data-driven fuzzy partitioning to better accommodate heterogeneous data sources. Exploring alternative evolutionary strategies or hybrid optimization schemes could also further refine the feature-selection process. These directions represent opportunities for expansion rather than shortcomings of the current system, which already establishes a high-performing and scalable foundation for clinically meaningful diabetes prediction.